\pdfoutput=1
\documentclass[11pt,a4paper]{article}
\usepackage[hyperref]{emnlp2020}
\usepackage{times}
\usepackage{latexsym}
\usepackage{soul}
\usepackage{url}

\usepackage{graphicx}
\usepackage{amsmath}
\usepackage{pifont}
\usepackage{amsfonts}
\usepackage{amssymb}
\usepackage{amsthm}
\usepackage{booktabs}
\usepackage{algorithm}
\usepackage{algorithmic}
\usepackage{xparse}
\usepackage{boldline}
\usepackage{tikz}
\usetikzlibrary{shapes,arrows}
\usepackage{caption}
\usepackage{subcaption}
\usepackage{tikz-dependency}
\usepackage{wrapfig}
\usepackage{pgfplots}
\usepackage{filecontents}
\usepackage[T1]{fontenc}
\usepackage{subfiles}
\usepackage{readarray}
\usepackage{xcolor}
\usepackage{import}
\usepackage{multirow}
\usepackage{hhline}
\usepackage{mathtools}
\usepackage{enumitem}
\usepackage{microtype}

\aclfinalcopy 
\def\aclpaperid{2971} 

\newcommand{\xvec}{\mathbf{x}}
\newcommand{\yvec}{\mathbf{y}}

\newcommand{\evec}{\mathbf{e}}

\newcommand{\rvec}{\mathbf{r}}

\newcommand{\mcL}{\mathcal{L}}

\newcommand{\cmark}{\textcolor{blue}{\ding{51}}}
\newcommand{\xmark}{\textcolor{red}{\ding{55}}}

\title{More Embeddings, Better Sequence Labelers?}

\author{\parbox{\linewidth}{\centering Xinyu Wang$^{\diamond\ddagger}$, Yong Jiang$^{\dagger}$\textsuperscript{$\ast$}, Nguyen Bach$^{\dagger}$, \\ Tao Wang$^{\dagger}$, Zhongqiang Huang$^{\dagger}$, Fei Huang$^{\dagger}$,  Kewei Tu$^{\diamond}$\thanks{\hspace{1mm} Yong Jiang and Kewei Tu are the corresponding authors. $^{\ddagger}$: This work was conducted when Xinyu Wang was interning at Alibaba DAMO Academy.}} \\
 $^\diamond$School of Information Science and Technology, ShanghaiTech University \\
 $^{\diamond}$Shanghai Engineering Research Center of Intelligent Vision and Imaging \\
 $^{\diamond}$University of Chinese Academy of Sciences \\
 $^\dagger$DAMO Academy, Alibaba Group \\
  {\tt \{wangxy1,tukw\}@shanghaitech.edu.cn} \\
  {\tt \{yongjiang.jy,nguyen.bach\}@alibaba-inc.com} \\
  {\tt \{leeo.wangt,z.huang,f.huang\}@alibaba-inc.com} \\
 }

\date{}

\begin{document}
\maketitle
\begin{abstract}
Recent work proposes a family of contextual embeddings that significantly improves the accuracy of sequence labelers over non-contextual embeddings. However, there is no definite conclusion on whether we can build better sequence labelers by combining different kinds of embeddings in various settings. 
In this paper, we conduct extensive experiments on 3 tasks over 18 datasets and 8 languages to study the accuracy of sequence labeling with various embedding concatenations and make three observations: (1) concatenating more embedding variants leads to better accuracy in rich-resource and cross-domain settings and some conditions of low-resource settings; (2) concatenating contextual sub-word embeddings with contextual character embeddings hurts the accuracy in extremely low-resource settings; (3) based on the conclusion of (1), concatenating additional similar contextual embeddings cannot lead to further improvements. We hope these conclusions can help people build stronger sequence labelers in various settings.

\end{abstract}

\section{Introduction}
\label{sec:introduction}
In recent years, sequence labelers equipped with contextual embeddings have achieved significant accuracy improvement \cite{peters-etal-2018-deep,akbik-etal-2018-contextual,devlin-etal-2019-bert,martin2019camembert} over approaches that use static non-contextual word embeddings \cite{mikolov2013distributed} and character embeddings \cite{santos2014learning}.
Different types of embeddings have different inductive biases to guide the learning process. However, little work has been done to study how to concatenate these contextual embeddings and non-contextual embeddings to build better sequence labelers in multilingual, low-resource, or cross-domain settings over various sequence labeling tasks. In this paper, we empirically investigate the effectiveness of concatenating various kinds of embeddings for multilingual sequence labeling and try to answer the following questions:
\begin{enumerate}
    \item In rich-resources settings, does combining different kinds of contextual embeddings result in a better sequence labeler? Are non-contextual embeddings helpful when the models are equipped with contextual embeddings?
    \item When we train models in low-resource and cross-domain settings, do the conclusions from the rich-resource settings still hold? 
    \item Can sequence labelers automatically learn the importance of each kind of embeddings when they are concatenated?
\end{enumerate}


\section{Model Architecture}
\subsection{Sequence Labeling}
We use the BiLSTM structure for all the sequence labeling tasks, which is one of the most popular approaches to sequence labeling \cite{huang2015bidirectional,ma-hovy-2016-end}. Given a $n$ word sentence $\xvec = \{x_1, \cdots, x_n\}$ and $L$ kinds of embeddings, we feed the sentence to generate the $l$-th kind of word embeddings $\{\evec^l_1, \cdots, \evec^l_n\}$:
\begin{align}
     \evec_i^l &= \text{embed}^l (\xvec) \nonumber
\end{align}
We concatenate these embeddings to generate the word representations $\{\rvec_1, \cdots, \rvec_n\}$ as the input of the BiLSTM layer:
\begin{align}
     \rvec_i &= \evec_i^1 \oplus \dots \oplus \evec_i^L \nonumber
\end{align}
where $\oplus$ represents the vector concatenation operation. We feed the word representations into a single-layer BiLSTM to generate the contextual hidden layer of each word. Then we use either a Softmax layer (the MaxEnt approach) or a Conditional Random Field layer (the CRF approach) \cite{10.5555/645530.655813,lample-etal-2016-neural,ma-hovy-2016-end} fed with the hidden layers to generate the conditional probability $p(\yvec|\xvec)$. Given the corresponding sequence of gold labels $\yvec^* = \{y_1^*, \cdots, y_n^*\}$ for the input sentence, the loss function for a model with parameters $\theta$ is:
\begin{displaymath}
\mcL_{\theta} = - \log p(\yvec^*|\xvec;\theta)
\end{displaymath}

\subsection{Embeddings}
There are mainly four kinds of embeddings that have been proved effective on the sequence labeling task: contextual sub-word embeddings, contextual character embeddings, non-contextual word embeddings and non-contextual character embeddings\footnote{We do not use contextual word embeddings such as ELMo \cite{peters-etal-2018-deep} since \citet{akbik-etal-2018-contextual} showed that concatenating Flair embeddings with ELMo embeddings cannot further improve the accuracy.}.
As we conduct our experiments in multilingual settings, we need to select suitable embeddings from each category for the concatenation. 

\paragraph{Contextual Sub-word Embeddings (CSEs)} 
\textbf{CSEs} such as OpenAI GPT \cite{radford2018improving} and BERT \cite{devlin-etal-2019-bert} are based on transformer \cite{vaswani2017attention} and use WordPiece embeddings \cite{sennrich-etal-2016-neural,wu2016google} as input.
Much research has focused on improving BERT model's performance such as better masking strategy \cite{liu2019roberta} and cross-lingual training \cite{conneau2019cross}. 
Since we focus on the multilingual settings of sequence labeling tasks, we use multilingual BERT (M-BERT), as recent researches shows its strong generalizability over various languages and tasks \cite{pires-etal-2019-multilingual,karthikeyan2020cross}.

\paragraph{Contextual Character Embeddings (CCEs)} 
\citet{liu2018empower} proposed a character language model by applying the BiLSTM over the sentence and trained jointly with the sequence labeling task. (Pooled) Contextual string embeddings (Flair) \cite{akbik-etal-2018-contextual,akbik-etal-2019-pooled} are pretrained on a large amount of unlabeled data and result in significant improvements for sequence labeling tasks. We use the Flair embeddings due to their high accuracy for sequence labeling task\footnote{We do not use the pooled version of Flair due to its slower speed in training.}.

\paragraph{Non-contextual Word Embeddings (NWEs)} The most common approach to the \textbf{NWEs} is Word2vec \cite{mikolov2013distributed}, which is a skip-gram model learning word representations by predicting neighboring words. Based on this approach, GloVe \cite{pennington2014glove} creates a co-occurrence matrix for global information and fastText \cite{bojanowski2017enriching} represents each word as an n-gram of characters. We use fastText in our experiments as there are pretrained embeddings for 294 languages.

\paragraph{Non-contextual Character Embeddings (NCEs)} 
Using character information to represent the embeddings of word is proposed by \citet{santos2014learning} with a lot of following work using a CNN structure to encode character representation \cite{dos-santos-guimaraes-2015-boosting,chiu-nichols-2016-named,ma-hovy-2016-end}. \citet{lample-etal-2016-neural} utilized BiLSTM on the character sequence of each word. We follow this approach as it usually results in better accuracy \cite{yang-etal-2018-design}.

\section{Experiments and Results}
For simplicity, we use \textbf{M} to represent M-BERT embeddings, \textbf{F} to represent Flair embeddings, \textbf{W} to represent fastText embeddings, \textbf{C} to represent non-contextual character embeddings, \textbf{All} to represents the concatenation of all types of embeddings and the operator ``+'' to represent the concatenation operation. We use the MaxEnt approach for all experiments\footnote{We find that the observations from the MaxEnt experiments do not change in all experiments with the CRF approach.}. Due to the space limit, some detailed experiment settings, extra experiments and discussions are included in the appendix. 

\subsection{Settings}
\paragraph{Datasets}
We use datasets from three multilingual sequence labeling tasks over 8 languages in our experiments: WikiAnn NER datasets \cite{pan-etal-2017-cross}, UD Part-Of-Speech (POS) tagging datasets \cite{nivre-etal-2016-universal}, and CoNLL 2003 chunking datasets \cite{tjong-kim-sang-de-meulder-2003-introduction}. We use language-specific fastText and Flair embeddings depending on the dataset. 

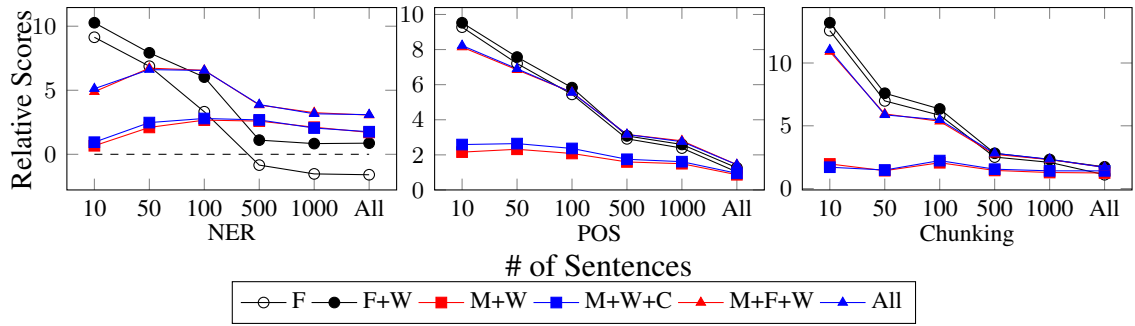
\begin{figure*}[ht]
\centering
\begin{tikzpicture}
    \node at (7.0,-1.0) {\large \# of Sentences};
    \node [rotate=90] at (-0.6,1.2) {Relative Scores};
    \begin{axis}[
        name=ner,
        width=0.37\textwidth,
        height=0.25\textwidth,
        xlabel=\small NER,
        legend columns=6, 
        legend pos=north west,
        legend style={font=\small,at={(0.5,-0.5)}},
        tick label style={font=\small},
        xticklabels={10,50,100,500,1000,All},
        xtick={1,2,3,4,5,6},
        xlabel style={yshift=0.2cm},
        ]
        \addplot[black,mark=o] table[x=sentences,y=flair] {wikiner.dat};
        \addplot[black,mark=*] table[x=sentences,y=word+flair] {wikiner.dat};
        \addplot[red,mark=square*] table[x=sentences,y=mbert+word] {wikiner.dat};
        \addplot[blue,mark=square*] table[x=sentences,y=mbert+word+char] {wikiner.dat};
        \addplot[red,mark=triangle*] table[x=sentences,y=mbert+word+flair] {wikiner.dat};
        \addplot[blue,mark=triangle*] table[x=sentences,y=mbert+word+flair+char] {wikiner.dat};
        \addplot[black,dashed] table[x=sentences,y=mbert] {bert.dat};
        \legend{F,F+W,M+W,M+W+C,M+F+W,All}
    \end{axis}
    \begin{axis}[
        at={(ner.south east)},
        xshift=0.5cm,
        name=pos,
        width=0.37\textwidth,
        height=0.25\textwidth,
        xlabel=\small POS,
        legend columns=4, 
        legend pos=north west,
        legend style={font=\tiny,at={(0.5,-0.5)}},
        tick label style={font=\small},
        xticklabels={10,50,100,500,1000,All},
        xtick={1,2,3,4,5,6},
        xlabel style={yshift=0.2cm},
        ]
        \addplot[black,mark=o] table[x=sentences,y=flair] {pos.dat};
        \addplot[black,mark=*] table[x=sentences,y=word+flair] {pos.dat};
        \addplot[red,mark=square*] table[x=sentences,y=mbert+word] {pos.dat};
        \addplot[blue,mark=square*] table[x=sentences,y=mbert+word+char] {pos.dat};
        \addplot[red,mark=triangle*] table[x=sentences,y=mbert+word+flair] {pos.dat};
        \addplot[blue,mark=triangle*] table[x=sentences,y=mbert+word+flair+char] {pos.dat};
        table[x=sentences,y=mbert] {bert.dat};
    \end{axis}
    \begin{axis}[
        at={(pos.south east)},
        xshift=0.5cm,
        name=chunk,
        width=0.37\textwidth,
        height=0.25\textwidth,
        xlabel=\small Chunking,
        legend columns=4, 
        legend pos=north west,
        legend style={font=\small,at={(0.15,-0.5)}},
        xticklabels={10,50,100,500,1000,All},
        tick label style={font=\small},
        xtick={1,2,3,4,5,6},
        xlabel style={yshift=0.2cm},
        ]
        \addplot[black,mark=o] table[x=sentences,y=flair] {chunk.dat};
        \addplot[black,mark=*] table[x=sentences,y=word+flair] {chunk.dat};
        \addplot[red,mark=square*] table[x=sentences,y=mbert+word] {chunk.dat};
        \addplot[blue,mark=square*] table[x=sentences,y=mbert+word+char] {chunk.dat};
        \addplot[red,mark=triangle*] table[x=sentences,y=mbert+word+flair] {chunk.dat};
        \addplot[blue,mark=triangle*] table[x=sentences,y=mbert+word+flair+char] {chunk.dat};
        table[x=sentences,y=mbert] {bert.dat};
    \end{axis}
\end{tikzpicture}
\caption{Relative score improvements against models with M-BERT embeddings for three tasks.}
\label{fig:lowres}
\end{figure*}

\paragraph{Embedding Concatenation} Since experimenting on all 15 concatenation combinations of the four embeddings is not essential for evaluating the effectiveness of each kind of embeddings, we experiment on the following 7 concatenations: \textbf{F}, \textbf{F+W}, \textbf{M}, \textbf{M+W}, \textbf{M+W+C}, \textbf{M+F+W}, \textbf{All}. Through these concatenations, we can answer the following questions: (1) whether \textbf{NWEs} are still helpful (\textbf{F} vs. \textbf{F+W} and \textbf{M} vs. \textbf{M+W}); (2) whether \textbf{NCEs} are still helpful (\textbf{M+W} vs. \textbf{M+W+C} and \textbf{M+F+W} vs. \textbf{All}); (3) whether concatenating different contextual embeddings results in a better sequence labeler (\textbf{F+W} vs. \textbf{M+F+W} and \textbf{M+W} vs. \textbf{M+F+W}); (4) which one is the best concatenation.



\begin{table}[t!]
\setlength\tabcolsep{4pt}
\centering
\small
\begin{tabular}{l|cccc||ccc|c}
\hlineB{4}
& \multicolumn{4}{c||}{\bf \textsc{Embeddings}} & \multicolumn{4}{c}{\bf \textsc{Tasks}} \\ \hhline{~|----||----}
& \textbf{M} & \textbf{F} & \textbf{W} & \textbf{C} &  {\bf\textsc{NER}}  & {\bf\textsc{POS}}  & {\bf\textsc{Chunk}} & {\bf\textsc{Avg.}} \\ 
\hline\hline
1. & \xmark & \cmark & \xmark & \xmark & 82.1 & 96.3 & 92.3 & 90.2 \\
2. & \xmark & \cmark & \cmark & \xmark & 84.6 & 96.5 & \textbf{92.9} & 91.4 \\
3. & \cmark & \xmark & \xmark & \xmark & 83.8 & 95.3 & 91.3 & 90.1 \\
4. & \cmark & \xmark & \cmark & \xmark & 85.5 & 96.1 & 92.5 & 91.4 \\
5. & \cmark & \xmark & \cmark & \cmark & 85.5 & 96.2 & 92.6 & 91.5 \\
6. & \cmark & \cmark & \cmark & \xmark & \textbf{86.8} & \textbf{96.7} & \textbf{92.9} & \textbf{92.1} \\
7. & \cmark & \cmark & \cmark & \cmark & \textbf{86.8} & \textbf{96.7} & \textbf{92.9} & \textbf{92.1} \\
\hlineB{4}
\end{tabular}
\caption{Averaged F1 scores over languages for each task with different embedding concatenations.}
\label{tab:main_results}
\end{table}

\subsection{Rich-resource and Low-resource Settings}
How to build better sequence labelers through embedding concatenations in both rich-resource and low-resource settings is the most important concern for users. We report the results of various concatenations of embeddings for the tasks in Table \ref{tab:main_results} for rich-resource settings and in Figure \ref{fig:lowres} for low-resource settings. From the results, we have the following observations.

\noindent\textbf{Observation \#1. Concatenating more embedding variants results in better sequence labelers:} In rich-resource settings, concatenating more embedding variants (\textbf{M+F+W} and \textbf{All}) results in best scores in most of the cases, which indicates that the inductive biases in various kind of embeddings are helpful to train a better sequence labeler.
In low-resource settings, \textbf{M+F+W} and \textbf{All} performs inferior to the \textbf{F+W} when the number of sentences are lower than 100. However, when the training set gets larger, the gap between these concatenations becomes smaller and reverses when the training set becomes larger than 100 for NER and POS tagging and the gap also disappears for Chunking. A possible reason is that using \textbf{CSEs} makes the model sample inefficient so that \textbf{CSEs} requires more training samples to improve accuracy than \textbf{CCEs}. The observation suggests that concatenating more embedding variants performs better if the training set is not extremely small.

\noindent\textbf{Observation \#2. \textbf{NCEs} become less effective when concatenated with \textbf{CSEs} and \textbf{CCEs}:} Concatenating \textbf{NCEs} with \textbf{CSEs} only marginally improves the accuracy. There is almost no improvement when concatenated with both \textbf{CSEs} and \textbf{CCEs} but the \textbf{NCEs} does not hurt the accuracy as well. A possible reason is that the \textbf{CSEs} and \textbf{CCEs} largely contain the information in \textbf{NCEs}\footnote{The observation is consistent with the observation of \citet{akbik-etal-2018-contextual}, but we experimented on more languages and tasks with the M-BERT embeddings.}.

\noindent\textbf{Observation \#3. \textbf{NWEs} are significantly helpful on top of contextual embeddings:} Although models based on contextual embeddings have proved to be stronger than models based on \textbf{NWEs} for sequence labeling, concatenating \textbf{NWEs} with contextual embeddings can still improve the accuracy significantly. The results imply that the contextual embeddings contain more contextual information over the input but lack static word information.

From these observations, we find that in most of rich-resource and low-resource settings, concatenating all embeddings variants or all embeddings variants except \textbf{NCEs} is the simplest choice for a better sequence labeler.

\begin{table}[t]
\small
\centering
\setlength\tabcolsep{2.5pt}
\begin{tabular}{l||ccccccc}
\hlineB{4}
 & \textbf{F} & \textbf{F+W} & \textbf{M} & \textbf{M+W} & \textbf{M+W+C} & \textbf{M+F+W} & \textbf{All} \\
 \hline
\textsc{\textbf{ Avg.}} & 46.3 & 48.6 & 47.4 & 48.4 & 48.7 & 49.9 & \textbf{50.4}\\
\hlineB{4}
\end{tabular}
\caption{Cross-domain transfer from the Wikipedia domain to the news domain on the NER task.}
\label{tab:cross_domain}
\end{table}

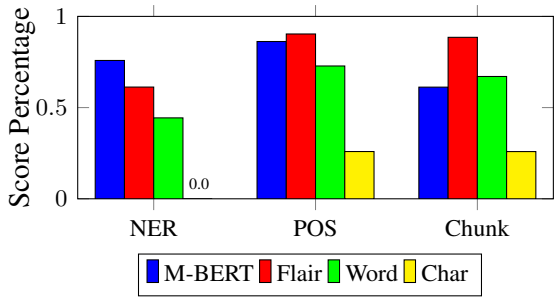
\begin{figure}[t]
\centering
\begin{tikzpicture}

\begin{axis}[
    ybar=0pt,
    bar width = {1em},
    width=0.49\textwidth,
    height=0.25\textwidth,
    enlarge x limits={abs=1cm},
    symbolic x coords={NER,POS,Chunk},
    xticklabel style={
         align=center, 
         font=\small, 
      },
    yticklabel style={font=\small},
    ylabel = {Score Percentage},
    ylabel style = {yshift=-0.5cm},
    legend style={font=\small,at={(0.85,-0.3)}},
    legend columns=4, 
    xtick={NER,POS,Chunk},
    legend image code/.code={%
      \draw[#1] (0cm,-0.1cm) rectangle (0.15cm,0.2cm);
    },
    ytick={0,0.5,1},
    ymin=0,
    ymax=1,
    ]
    \addplot[ybar,fill=blue] table [x=type, y=MBERT] {mask_embed.dat};
    \addplot[ybar,fill=red] table [x=type, y=Flair] {mask_embed.dat};
    \addplot[ybar,fill=green] table [x=type, y=Word] {mask_embed.dat};
    \addplot[ybar,fill=yellow] table [x=type, y=char] {mask_embed.dat};
    \legend{M-BERT,Flair,Word,Char};
\end{axis}
\node at (1.6,0.2) {\tiny 0.0};
\end{tikzpicture}
\caption{Importance of each embedding over the concatenation of \textbf{All} embeddings. The score percentage represents the average score preserving only one kind of embeddings divided by the score without masking.}
\label{fig:masking}
\end{figure}

\subsection{Cross-domain Settings}
Another concern for users is that we want to build better sequence labelers not only in in-domain settings but in out-of-domain settings as well. We conduct experiments in cross-domain settings to show how the embedding concatenations impact the accuracy when the distribution of training data and test data are different. We evaluate our Wikipedia NER models on CoNLL 2002/2003 NER \cite{tjong-kim-sang-2002-introduction,tjong-kim-sang-de-meulder-2003-introduction} datasets from the news domain. The results (Table \ref{tab:cross_domain}) are almost consistent with rich-resource settings, suggesting that concatenating more embedding variants results in better sequence labelers.

\subsection{Importance of Embeddings}
\label{sec:importance}
To study the effectiveness of concatenating embeddings from another perspective, we preserve only one kind of embedding in \textbf{All} and mask out the other embeddings as 0 to study how the models rely upon each kind of embeddings. To avoid the impact of embedding dimensions, we train the model by linearly projecting each kind of embeddings into the same dimension of 4096. The results (Figure \ref{fig:masking}) show that the accuracy of preserved embeddings has a positive correlation with the results in Table \ref{tab:main_results}. For example, \textbf{M} gets higher accuracy than other embeddings in NER and Table \ref{tab:main_results} also shows that the model with \textbf{F} performs inferior to the model with \textbf{M} only. The models with concatenated embeddings almost do not rely on \textbf{NCEs} and relies mostly on \textbf{CSEs} or \textbf{CCEs} depending on the task. These results show that models with concatenated embeddings can extract helpful information from each kind of embeddings to improve accuracy.

\begin{table}[t!]
\small
\centering
\setlength\tabcolsep{5pt}
\begin{tabular}{cccccc||ccc}
\hlineB{4}
\multicolumn{6}{c||}{\bf \textsc{Embeddings}} & \multicolumn{3}{c}{\bf \textsc{Tasks}} \\ 
\hline
\textbf{M} & \textbf{F} & \textbf{W} & \textbf{C} & \textbf{B} &  \textbf{MF} & {\bf\textsc{NER}}  & {\bf\textsc{POS}}  & {\bf\textsc{Chunk}}\\ 
\hline
\multicolumn{9}{c}{\bf +En-BERT (English)}\\
\hline
\cmark & \cmark & \cmark & \cmark & \xmark & \xmark & 81.8 & 97.0 & 91.6 \\
\xmark & \cmark & \cmark & \cmark & \cmark & \xmark & 80.5 & \textbf{97.2} & \textbf{91.8} \\
\cmark & \cmark & \cmark & \cmark & \cmark & \xmark & \textbf{82.1} & \textbf{97.2} & 91.6 \\
\hline
\multicolumn{9}{c}{\bf +M-Flair (All languages)}\\
\hline
\cmark & \cmark & \cmark & \cmark & \xmark & \xmark & \textbf{86.8} & \textbf{96.7} & \textbf{92.9} \\
\cmark & \xmark & \cmark & \cmark & \xmark & \cmark & 86.1 & 96.5 & 92.8 \\
\cmark & \cmark & \cmark & \cmark & \xmark & \cmark & \textbf{86.8} & \textbf{96.7} & \textbf{92.9} \\
\hlineB{4}
\end{tabular}
\caption{Comparisons of the effectiveness for additionally concatenating the same category of embeddings. \textbf{B} represents the En-BERT embeddings and \textbf{MF} represents the M-Flair embeddings.}
\label{tab:same_embed}
\end{table}

\subsection{On Concatenating Similar Embeddings}
Since concatenating more embeddings variants results in better sequence labelers, we additionally concatenate multilingual Flair embeddings (M-Flair) or English BERT embeddings (En-BERT) with \textbf{All} embeddings to show whether concatenating the same category of embeddings can further improve the accuracy. We evaluate the addition of En-BERT on English and M-Flair on all languages in each task. The results are shown in Table \ref{tab:same_embed}. It can be seen that additionally concatenating the same category of embeddings does not further improve the accuracy in most cases except for concatenating En-BERT on English WikiAnn NER. A possible reason is that the BERT models are trained on the same domain as WikiAnn and hence the inductive biases of BERT embeddings help improve the accuracy.

We also find that concatenating En-BERT with \textbf{All} only improves the accuracy of WikiAnn English NER. We think the possible reason for the improvement is that the BERT and the training data have the same domain of Wikipedia. We conduct the same concatenation on the CoNLL English NER dataset for comparison. The results in Table \ref{tab:en_m_bert} show that concatenating En-BERT with \textbf{All} does not further improve the accuracy on CoNLL English NER.

\begin{table}[t!]
\setlength\tabcolsep{4pt}
\small
\centering
\begin{tabular}{l|ccc}
\hlineB{4}
{\bf \textsc{Embeddings}} & \multicolumn{3}{c}{\bf \textsc{Tasks}} \\  
 \hline
 & {\bf\textsc{NER}}  & {\bf\textsc{POS}}  & {\bf\textsc{Chunk}} \\
 \hline
\textbf{F+W} & 32.7 & 81.7 & 78.2 \\
\textbf{F+W}+Proj. & \textbf{33.2} & \textbf{82.3} & \textbf{79.0} \\
\textbf{All} & 27.5 & 80.4 & 76.1 \\
\hlineB{4}
\end{tabular}
\caption{Comparisons of \textbf{F+W}, \textbf{All}, and \textbf{F+W+proj} (\textbf{F+W} with linearly projecting the hidden size into the hidden size of \textbf{All}) in three tasks with 10-sentence low-resource settings. The accuracy is averaged over tasks.}
\label{tab:proj}
\end{table}

\begin{table}[t!]
\setlength\tabcolsep{4pt}
\small
\centering
\begin{tabular}{l|ccccc||ccc}
\hlineB{4}
& \multicolumn{5}{c||}{\bf \textsc{Embeddings}} & \multicolumn{3}{c}{\bf \textsc{Tasks}} \\  
\hhline{~|-----||---}
& \textbf{M} & \textbf{F} & \textbf{W} & \textbf{C} & \textbf{B}  & {\bf\textsc{NER}}  & {\bf\textsc{POS}}  & {\bf\textsc{Chunk}} \\
\hline\hline
\multicolumn{9}{c}{\bf \textsc{Low-Resource: 10 Sentences}}\\
\hline
1. & \xmark & \cmark & \cmark & \xmark & \xmark & \textbf{35.5}$\pm1.4$ & \textbf{80.2}$\pm0.1$ & \textbf{73.3}$\pm0.6$ \\
2. & \cmark & \cmark & \cmark & \cmark & \xmark & 25.4$\pm0.8$ & 77.9$\pm0.2$ & 70.8$\pm0.5$ \\
3. & \xmark & \cmark & \cmark & \cmark & \cmark & 29.3$\pm0.8$ & 79.6$\pm0.2$ & 67.9$\pm0.5$ \\
\hline\hline
\multicolumn{9}{c}{\bf \textsc{Low-Resource: 50 Sentences}}\\
\hline
1. & \xmark & \cmark & \cmark & \xmark & \xmark & \textbf{48.6}$\pm0.3$ & 88.8$\pm0.0$ & \textbf{82.2}$\pm0.0$ \\
2. & \cmark & \cmark & \cmark & \cmark & \xmark & 48.5$\pm0.4$ & 87.5$\pm0.1$ & 80.3$\pm0.3$ \\
3. & \xmark & \cmark & \cmark & \cmark & \cmark & 43.4$\pm0.9$ & \textbf{88.9}$\pm0.0$ & 78.8$\pm0.1$ \\
\hline\hline
\multicolumn{9}{c}{\bf \textsc{Low-Resource: 100 Sentences}}\\
\hline
1. & \xmark & \cmark & \cmark & \xmark & \xmark & 54.8$\pm0.5$ & 90.6$\pm0.1$ & \textbf{83.7}$\pm0.0$ \\
2. & \cmark & \cmark & \cmark & \cmark & \xmark & \textbf{56.8}$\pm0.1$ & 90.3$\pm0.0$ & 82.4$\pm0.0$ \\
3. & \xmark & \cmark & \cmark & \cmark & \cmark & 50.2$\pm0.5$ & \textbf{91.4}$\pm0.1$ & 82.9$\pm0.1$ \\
\hline\hline
\multicolumn{9}{c}{\bf \textsc{Low-Resource: 500 Sentences}}\\
\hline
1. & \xmark & \cmark & \cmark & \xmark & \xmark & 68.3$\pm0.2$ & 92.8$\pm0.0$ & 86.8$\pm0.0$ \\
2. & \cmark & \cmark & \cmark & \cmark & \xmark & \textbf{69.1}$\pm0.2$ & 93.0$\pm0.1$ & 86.7$\pm0.1$ \\
3. & \xmark & \cmark & \cmark & \cmark & \cmark & 67.3$\pm0.1$ & \textbf{93.9}$\pm0.1$ & \textbf{86.9}$\pm0.0$ \\
\hline\hline
\multicolumn{9}{c}{\bf \textsc{Low-Resource: 1000 Sentences}}\\
\hline
1. & \xmark & \cmark & \cmark & \xmark & \xmark & 72.0$\pm0.1$ & 94.0$\pm0.1$ & 87.1$\pm0.1$ \\
2. & \cmark & \cmark & \cmark & \cmark & \xmark & \textbf{75.2}$\pm0.3$ & 94.4$\pm0.1$ & 87.1$\pm0.2$ \\
3. & \xmark & \cmark & \cmark & \cmark & \cmark & 70.8$\pm0.1$ & \textbf{95.0}$\pm0.0$ & \textbf{87.6}$\pm0.1$ \\
\hline\hline
\multicolumn{9}{c}{\bf \textsc{Rich-Resource}}\\
\hline
1. & \xmark & \cmark & \cmark & \xmark & \xmark & 79.9$\pm0.3$ & 96.7$\pm0.0$ & 91.7$\pm0.1$ \\
2. & \cmark & \cmark & \cmark & \cmark & \xmark & \textbf{81.7}$\pm0.2$ & 97.0$\pm0.1$ & 91.6$\pm0.1$ \\
3. & \xmark & \cmark & \cmark & \cmark & \cmark & 80.5$\pm0.2$ & \textbf{97.2}$\pm0.0$ & \textbf{91.8}$\pm0.1$\\
\hlineB{4}
\end{tabular}
\caption{Comparisons of using English BERT instead of M-BERT in English datasets. \textbf{B} represents the En-BERT embeddings. We also provide the concatenation of Flair and pretrained word embeddings for reference.}
\label{tab:bert}
\end{table}

\begin{table}[t!]
\setlength\tabcolsep{4pt}
\small
\centering
\begin{tabular}{l|ccc}
\hlineB{4}
{\bf \textsc{Embeddings}} & \multicolumn{3}{c}{\bf \textsc{Tasks}} \\  
 \hline
 & {\bf\textsc{NER}}  & {\bf\textsc{POS}}  & {\bf\textsc{Chunk}} \\
 \hline
\textbf{All} & \textbf{86.8} & \textbf{96.7} & \textbf{92.9} \\
\textbf{All}+50d Proj. & 83.8 & 96.3 & 92.0 \\
\textbf{All}+1024d Proj. & 84.8 & 96.5 & 92.2 \\
\textbf{All}+4096d Proj. & 85.1 & 96.5 & 92.2 \\
\hlineB{4}
\end{tabular}
\caption{Comparisons of \textbf{All} with different linear projection size in three tasks with rich-resource settings. The accuracy is averaged over tasks.}
\label{tab:eachproj}
\end{table}

\subsection{English BERT vs. M-BERT}
We use English BERT embeddings instead of M-BERT embeddings to see whether the language-specific \textbf{CSEs} impact the observations. The results (Table \ref{tab:bert}) show that our observations do not change in both rich-resource and low-resource settings. Using a language-specific BERT embedding can even get better sequence labelers for the POS tagging and chunking tasks in rich-resource settings.

\subsection{Hidden Sizes and Accuracy}
In low-resource settings with 10 sentences, we find that models with \textbf{All} perform inferior to the models with \textbf{F+W}. One possible concern is that whether the larger hidden size of \textbf{All} introduces more parameters in the model and makes the model over-fits the training set. We linearly project the hidden size of \textbf{F+W} (4396) to the same hidden size as \textbf{All} (5214). Table \ref{tab:proj} shows that with linear projection, \textbf{F+W} performs even better. Therefore, the cause for over-fitting is not the inferior accuracy of \textbf{All} but possibly the sample inefficiency for \textbf{CSEs}. 

Another concern is whether we can project each embedding to a larger hidden size to improve the accuracy. Since we try a projection to 4096 for each embedding in \textbf{F+W+proj} (Section \ref{sec:importance}), we further project each embedding variants to see how the projection affect the accuracy in rich-resource settings. The results (Table \ref{tab:eachproj}) show that the linear projection for each embedding significantly decreases the accuracy of the models.

From the two experiments, we find that the hidden sizes of concatenated embeddings do not impact the observations.

\begin{table}[t!]
\setlength\tabcolsep{4pt}
\small
\centering
\begin{tabular}{ccccc||c}
\hlineB{4}
\multicolumn{5}{c||}{\bf \textsc{Embeddings}} & \multicolumn{1}{c}{\bf \textsc{Task}} \\  
\hhline{-----||-}
\textbf{M} & \textbf{F} & \textbf{W} & \textbf{C} & \textbf{B}  & {\bf\textsc{English NER}} \\
\hline
\cmark & \cmark & \cmark & \cmark & \xmark & \textbf{92.1}$\pm0.1$\\
\xmark & \cmark & \cmark & \cmark & \cmark & 92.0$\pm0.1$\\
\cmark & \cmark & \cmark & \cmark & \cmark & \textbf{92.1}$\pm0.1$\\
\hlineB{4}
\end{tabular}
\caption{Comparisons of concatenating En-BERT with \textbf{All} on CoNLL NER. \textbf{B} represents the En-BERT.}
\label{tab:en_m_bert}
\end{table}




\section{Conclusion}
In this paper, we analyze how to get a better sequence labeler by concatenating various kinds of embeddings. We make several empirical observations that we hope can guide future work to build better sequence labelers: 
(1) in most settings, concatenating more embedding variants leads to better results, while in extremely low-resource settings, only using \textbf{CSEs} and \textbf{NWEs} performs better; (2) \textbf{NCEs} become less effective when concatenated with contextual embeddings, while \textbf{NWEs} are still beneficial; (3) neural models can automatically learn which embeddings are beneficial to the task; (4) additionally concatenating similar contextual embeddings with the best concatenations from (1) cannot further improve the accuracy in most cases.

\section*{Acknowledgements}
This work was supported by the National Natural Science Foundation of China (61976139). This work also was supported by Alibaba Group through Alibaba Innovative Research Program. 
The authors wish to thank Chao Lou for his helpful comments and suggestions.

\bibliographystyle{acl_natbib}
\bibliography{anthology,emnlp2020}


\appendix

\section{Appendix}
In this appendix, we use ISO 639-1 codes\footnote{\url{https://en.wikipedia.org/wiki/List_of_ISO_639-1_codes}} to represent each language for simplification.
\subsection{Settings}
\paragraph{Datasets}
We use the following datasets for experiments:
\begin{itemize}
    \item \textbf{Named Entity Recognition (NER)}: We use \textbf{{\bf WikiAnn}}\footnote{\url{https://elisa-ie.github.io/wikiann/}} \cite{pan-etal-2017-cross} datasets and CoNLL 2002/2003 NER\footnote{\url{https://www.clips.uantwerpen.be/conll2003/ner/}} \cite{tjong-kim-sang-2002-introduction,tjong-kim-sang-de-meulder-2003-introduction} datasets  for experiments. The {\bf WikiAnn} datasets contain silver standard NER tags over 282 languages. We select 8 languages from the dataset. We randomly choose 5000 sentences from the dataset for each language except English with 12000 sentences. We split the dataset by 3:1:1 for training/development/test. We use the standard training/development/test split for the CoNLL NER experiments.
    \item \textbf{Part-Of-Speech (POS) tagging}: We use universal POS tagging annotations in the {\bf Universal Dependencies} (UD) \cite{nivre-etal-2016-universal} datasets\footnote{\url{https://lindat.mff.cuni.cz/repository/xmlui/handle/11234/1-2837}}. We choose one treebank for each language from the same 8 languages that are used in the WikiAnn experiments. The list of treebank are shown in Table \ref{tab:treebank}. We use the official train/development/test split for experiments.
    \item \textbf{Chunking}: We use the chunking datasets from the CoNLL 2003 shared task \cite{tjong-kim-sang-de-meulder-2003-introduction}, which contain two languages for chunking. We use the official train/development/test split for experiments.
\end{itemize}

\paragraph{Model Configuration and Running} For the embeddings, the hidden sizes for fastText and Flair embeddings\footnote{Details of Flair embeddings \url{https://github.com/flairNLP/flair/blob/master/resources/docs/embeddings/FLAIR_EMBEDDINGS.md}} are 300 and 4096, respectively. The dimension of character embeddings is set to 50\footnote{We did not observe further gains when increasing the dimension size.} following previous work \cite{lample-etal-2016-neural}. For M-BERT embeddings, we use the cased version that is trained on 104 languages for all datasets. We use the official release of \textit{bert-base-cased} model in the experiments for English BERT. The word embeddings are fine-tuned and character embeddings are trained for tasks while the Flair and BERT embeddings are fixed. Our codes are mainly based on the official release of Flair\footnote{\url{https://github.com/flairNLP/flair}} which is based on PyTorch v1.1.0 \cite{NEURIPS2019_9015}. We run our experiments on a GPU server with NVIDIA Tesla V100 GPU. For model training, we set the mini-batch size to 2,000 tokens for better GPU utilization. Following the official release of Flair, we use an SGD optimizer with a learning rate of 0.1 for training all models and set the hidden size of BiLSTM to 256. We anneal the learning rate by 0.5 if there is no improvement on the development sets for 10 and 100 epochs when training rich-resource and low-resource datasets respectively. We fix these hyper-parameters for all experiments because we find that tuning these hyper-parameters does not impact the observation and usually results in lower accuracy. We average over 5 runs for each experiment and report the macro-average score over all languages for each task. 

\paragraph{Pre-processing and Evaluation} We evaluate the NER and chunking by the F1 score and POS tagging by the accuracy. We use the evaluation script in the official release of Flair. We convert the BIO format into BIOES format for all NER and chunking datasets.


\subsection{Detailed Results}
\begin{figure*}[t!]
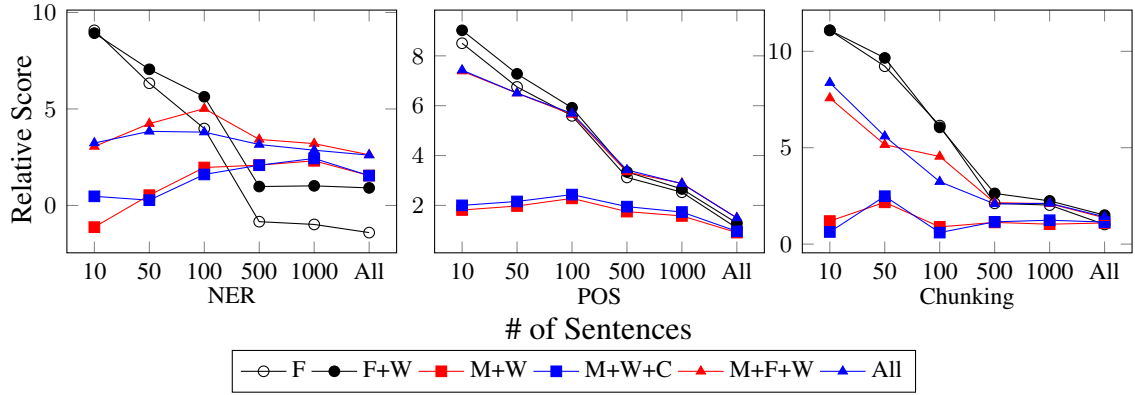

\centering
%
\caption{Relative score improvements against models with M-BERT embeddings for three tasks. Models are equipped with the CRF layer.}
\label{fig:crf_lowres}
\end{figure*}

For the models using the CRF layer, similar to the main paper, we plot our results in the rich-resource and low-resource settings in Figure \ref{fig:crf_lowres}. The figures have similar trends as the MaxEnt models, showing that output structures do not impact the observations.

Table \ref{tab:importance} shows the importance of each kind of embeddings for each language and task (Section 3.4 in the main paper). Table \ref{tab:ner}, \ref{tab:pos} and \ref{tab:chunk} show average scores over each language for each task in the rich-resource and low-resource settings (Section 3.2). Table \ref{tab:crf_ner}, \ref{tab:crf_pos} and \ref{tab:crf_chunk} show average scores over each language for each task in the rich-resource and low-resource settings. Table \ref{tab:app_cross_domain} shows the average scores for each language in our cross-domain experiments (Section 3.3). Table \ref{tab:mflair} show the detailed comparison for additionally concatenating M-Flair embeddings with \textbf{All} for all datasets (Section 3.5).

\begin{table}[t!]

\centering
\begin{tabular}{lr}
\hlineB{4}
Language & Treebank\\
\hline
ar & PADT\\
cs & FicTree\\
de & GSD\\
en & EWT\\
es & GSD\\
fr & Sequoia\\
nl & LassySmall\\
ta & TTB\\
\hlineB{4}
\end{tabular}
\caption{The list of treebank that we used in UD POS tagging.}
\label{tab:treebank}
\end{table}

\begin{table*}[ht]
\small
\centering
\begin{tabular}{l|cccc||cccc|c}
\hlineB{4}
& \multicolumn{4}{c||}{\bf \textsc{Embeddings}} & \multicolumn{5}{c}{\bf MaxEnt models on WikiAnn NER} \\ 
\hhline{~|----||-----}
 &  \textbf{M} & \textbf{F} & \textbf{W} & \textbf{C} & de & en & es & nl & avg \\
 \hline
1. & \xmark & \cmark & \xmark & \xmark & 41.0$\pm0.7$ & 46.9$\pm1.4$ & 47.4$\pm1.7$ & 49.8$\pm0.6$ & 46.3 \\
2. & \xmark & \cmark & \cmark & \xmark & 43.2$\pm0.6$ & 48.1$\pm1.0$ & \textbf{50.1}$\pm0.3$ & 52.9$\pm0.3$ & 48.6 \\
3. & \cmark & \xmark & \xmark & \xmark & 45.2$\pm0.9$ & 49.2$\pm0.7$ & 45.2$\pm2.8$ & 49.9$\pm0.7$ & 47.4 \\
4. & \cmark & \xmark & \cmark & \xmark & 43.5$\pm0.1$ & 50.9$\pm0.7$ & 46.9$\pm1.3$ & 52.5$\pm0.6$ & 48.4 \\
5. & \cmark & \xmark & \cmark & \cmark & 44.4$\pm1.2$ & 50.6$\pm0.6$ & 47.1$\pm0.1$ & 52.9$\pm0.6$ & 48.7 \\
6. & \cmark & \cmark & \cmark & \xmark & \textbf{46.7}$\pm0.4$ & 51.1$\pm0.7$ & 48.0$\pm2.1$ & 53.9$\pm0.9$ & 49.9 \\
7. & \cmark & \cmark & \cmark & \cmark & 46.6$\pm0.4$ & \textbf{52.3}$\pm0.5$ & 47.9$\pm0.3$ & \textbf{54.7}$\pm0.3$ & \textbf{50.4} \\
\hlineB{4}
\end{tabular}
\caption{Detailed results of cross-domain transfer from the Wikipedia domain to the news domain on the NER task. We use the ISO 639 language code to represent each language.}
\label{tab:app_cross_domain}
\end{table*}


\begin{table}[t!]
\setlength\tabcolsep{2.5pt}
\small
\centering
\begin{tabular}{l||cccc|c}
\hlineB{4}
 & M-BERT & Flair & Word & Char & \textbf{All} \\
\hline\hline
\multicolumn{6}{c}{WikiAnn NER}\\
\hline
ar & 53.0$\pm1.4$ & 43.7$\pm1.1$ & 44.9$\pm2.6$ & 0.0$\pm0.0$ & 82.6$\pm0.1$ \\
cs & 71.6$\pm0.4$ & 54.5$\pm2.4$ & 45.9$\pm2.1$ & 0.0$\pm0.0$ & 88.0$\pm0.1$ \\
de & 67.0$\pm0.8$ & 58.1$\pm2.5$ & 33.6$\pm1.9$ & 0.0$\pm0.0$ & 85.6$\pm0.2$ \\
en & 67.7$\pm0.9$ & 46.5$\pm0.9$ & 22.8$\pm0.5$ & 0.0$\pm0.0$ & 88.4$\pm0.7$ \\
es & 74.9$\pm1.0$ & 54.6$\pm2.5$ & 35.0$\pm2.0$ & 0.0$\pm0.0$ & 79.9$\pm0.6$ \\
fr & 75.9$\pm1.5$ & 48.0$\pm1.3$ & 35.2$\pm3.9$ & 0.0$\pm0.0$ & 84.2$\pm0.1$ \\
nl & 63.4$\pm3.7$ & 59.0$\pm1.6$ & 40.3$\pm0.9$ & 0.0$\pm0.0$ & 86.7$\pm0.4$ \\
ta & 41.3$\pm0.9$ & 51.5$\pm1.5$ & 43.5$\pm1.4$ & 0.0$\pm0.0$ & 83.4$\pm0.2$ \\
\hline
Avg. & 64.3 & 52.0 & 37.6 & 0.0 & 84.8 \\
\hline\hline
\multicolumn{6}{c}{UD POS tagging}\\
\hline
ar & 84.5$\pm1.7$ & 88.3$\pm0.7$ & 79.3$\pm0.6$ & 27.4$\pm0.9$ & 92.0$\pm0.6$ \\
cs & 84.0$\pm0.6$ & 90.8$\pm0.1$ & 68.5$\pm0.6$ & 25.6$\pm3.2$ & 96.5$\pm0.1$ \\
de & 78.3$\pm2.9$ & 88.3$\pm0.2$ & 71.3$\pm0.8$ & 26.3$\pm3.6$ & 98.8$\pm0.0$ \\
en & 85.1$\pm0.9$ & 85.8$\pm0.2$ & 65.3$\pm1.8$ & 32.5$\pm1.1$ & 97.2$\pm0.0$ \\
es & 83.2$\pm1.6$ & 92.4$\pm0.5$ & 80.9$\pm1.6$ & 20.4$\pm5.9$ & 96.6$\pm0.0$ \\
fr & 92.6$\pm0.7$ & 85.2$\pm0.6$ & 63.7$\pm0.2$ & 15.8$\pm3.0$ & 95.2$\pm0.0$ \\
nl & 80.9$\pm0.3$ & 89.9$\pm0.1$ & 70.9$\pm1.9$ & 18.9$\pm1.7$ & 98.7$\pm0.0$ \\
ta & 76.8$\pm1.8$ & 76.6$\pm0.3$ & 62.0$\pm0.0$ & 33.1$\pm4.8$ & 96.7$\pm0.0$ \\
\hline
Avg. & 83.2 & 87.2 & 70.2 & 25.0 & 96.5 \\
\hline\hline
\multicolumn{6}{c}{Chunking}\\
\hline
de & 66.3$\pm4.0$ & 90.2$\pm0.3$ & 52.9$\pm2.9$ & 28.0$\pm0.5$ & 93.5$\pm0.1$ \\
en & 46.6$\pm2.1$ & 73.0$\pm0.8$ & 70.7$\pm0.6$ & 19.8$\pm0.3$ & 90.8$\pm0.1$ \\
\hline
Avg. & 56.4 & 81.6 & 61.8 & 23.9 & 92.2 \\
\hlineB{4}
\end{tabular}
\caption{Detailed results on importance of embeddings for each language.}
\label{tab:importance}
\end{table}

\begin{table*}[t]
\setlength\tabcolsep{4pt}
\small
\centering
\begin{tabular}{l|cccc||cccccccc|c}
\hlineB{4}
& \multicolumn{4}{c||}{\bf \textsc{Embeddings}} & \multicolumn{9}{c}{\bf MaxEnt models on WikiAnn NER} \\  
\hhline{~|----||---------}
 &  \textbf{M} & \textbf{F} & \textbf{W} & \textbf{C}  & ar & cs & de & en & es & fr & nl & ta & Avg. \\
\hline\hline
\multicolumn{13}{c}{\bf \textsc{Low-Resource: 10 Sentences}}\\
\hline
1. & \xmark & \cmark & \xmark & \xmark & 33.6$\pm1.3$ & 36.9$\pm1.5$ & 25.5$\pm0.8$ & \textbf{35.8}$\pm1.2$ & \textbf{34.4}$\pm2.0$ & 39.4$\pm2.0$ & 26.2$\pm1.7$ & 20.6$\pm0.7$ & 31.5 \\
2. & \xmark & \cmark & \cmark & \xmark & \textbf{34.4}$\pm1.4$ & \textbf{40.5}$\pm1.4$ & \textbf{26.0}$\pm0.4$ & 35.5$\pm1.4$ & 33.9$\pm3.1$ & \textbf{41.6}$\pm0.2$ & 28.6$\pm1.9$ & \textbf{20.8}$\pm1.1$ & \textbf{32.7} \\
3. & \cmark & \xmark & \xmark & \xmark & 29.6$\pm2.4$ & 31.0$\pm0.9$ & 13.8$\pm1.9$ & 23.2$\pm1.7$ & 23.2$\pm1.5$ & 24.6$\pm2.3$ & 16.2$\pm1.5$ & 17.5$\pm0.8$ & 22.4 \\
4. & \cmark & \xmark & \cmark & \xmark & 28.8$\pm1.9$ & 32.2$\pm0.9$ & 15.7$\pm1.6$ & 22.1$\pm0.6$ & 23.1$\pm1.0$ & 24.2$\pm2.5$ & 20.0$\pm2.9$ & 18.6$\pm0.7$ & 23.1 \\
5. & \cmark & \xmark & \cmark & \cmark & 27.8$\pm0.6$ & 32.1$\pm1.5$ & 15.8$\pm2.2$ & 23.1$\pm0.5$ & 23.9$\pm1.3$ & 26.3$\pm2.3$ & 19.2$\pm1.8$ & 18.6$\pm0.4$ & 23.4 \\
\end{tabular}
\caption{Averaged F1 scores over 8 languages for WikiAnn NER.}
\label{tab:ner}
\end{table*}

\begin{table*}[t]
\setlength\tabcolsep{4pt}
\small
\centering
\begin{tabular}{l|cccc||cccccccc|c}
\hlineB{4}
& \multicolumn{4}{c||}{\bf \textsc{Embeddings}} & \multicolumn{9}{c}{\bf CRF models on WikiAnn NER} \\  
\hhline{~|----||---------}
 &  \textbf{M} & \textbf{F} & \textbf{W} & \textbf{C}  & ar & cs & de & en & es & fr & nl & ta & Avg. \\
\hline\hline
\multicolumn{13}{c}{\bf \textsc{Low-Resource: 10 Sentences}}\\
\hline
1. & \xmark & \cmark & \xmark & \xmark & 31.0$\pm5.2$ & 38.6$\pm1.1$ & 23.5$\pm0.7$ & \textbf{35.1}$\pm2.2$ & \textbf{37.6}$\pm3.6$ & 41.2$\pm2.2$ & 27.4$\pm3.4$ & 19.6$\pm1.1$ & \textbf{31.8} \\
2. & \xmark & \cmark & \cmark & \xmark & \textbf{32.0}$\pm3.3$ & \textbf{38.9}$\pm4.6$ & \textbf{24.1}$\pm1.3$ & 33.3$\pm0.5$ & 31.7$\pm2.0$ & \textbf{43.1}$\pm4.5$ & \textbf{30.4}$\pm3.9$ & 19.6$\pm3.7$ & 31.6 \\
3. & \cmark & \xmark & \xmark & \xmark & 31.5$\pm4.7$ & 30.0$\pm2.8$ & 12.5$\pm4.9$ & 23.1$\pm1.0$ & 23.5$\pm5.2$ & 24.7$\pm3.2$ & 18.3$\pm2.7$ & 17.8$\pm1.0$ & 22.7 \\
4. & \cmark & \xmark & \cmark & \xmark & 28.4$\pm3.2$ & 30.9$\pm1.9$ & 14.2$\pm1.3$ & 21.3$\pm2.9$ & 21.5$\pm1.9$ & 25.4$\pm3.4$ & 12.5$\pm8.0$ & 18.5$\pm1.0$ & 21.6 \\
5. & \cmark & \xmark & \cmark & \cmark & 30.9$\pm1.8$ & 32.8$\pm2.2$ & 11.0$\pm5.7$ & 19.7$\pm1.4$ & 24.7$\pm1.4$ & 28.1$\pm0.3$ & 19.3$\pm3.8$ & 18.8$\pm1.3$ & 23.2 \\
\end{tabular}
\caption{Averaged F1 scores over 8 languages for WikiAnn NER with the CRF layer.}
\label{tab:crf_ner}
\end{table*}

\begin{table*}[ht]
\setlength\tabcolsep{4pt}
\small
\centering
\begin{tabular}{l|cccc||cccccccc|c}
\hlineB{4}
& \multicolumn{4}{c||}{\bf \textsc{Embeddings}} & \multicolumn{9}{c}{\bf MaxEnt models on UD POS tagging} \\  
\hhline{~|----||---------}
 &  \textbf{M} & \textbf{F} & \textbf{W} & \textbf{C}  & ar & cs & de & en & es & fr & nl & ta & Avg. \\
\hline\hline
\multicolumn{13}{c}{\bf \textsc{Low-Resource: 10 Sentences}}\\
\hline
1. & \xmark & \cmark & \xmark & \xmark & 86.4$\pm0.3$ & \textbf{83.0}$\pm0.4$ & \textbf{83.4}$\pm0.6$ & 79.5$\pm0.1$ & 88.7$\pm0.1$ & 85.5$\pm0.2$ & \textbf{72.1}$\pm0.5$ & 72.5$\pm0.4$ & 81.4 \\
2. & \xmark & \cmark & \cmark & \xmark & \textbf{87.1}$\pm0.1$ & 82.5$\pm0.2$ & 83.1$\pm0.3$ & \textbf{80.2}$\pm0.1$ & \textbf{88.8}$\pm0.1$ & \textbf{86.2}$\pm0.2$ & \textbf{72.1}$\pm0.6$ & 73.2$\pm0.2$ & \textbf{81.7} \\
3. & \cmark & \xmark & \xmark & \xmark & 80.7$\pm0.7$ & 71.0$\pm1.3$ & 73.4$\pm1.1$ & 72.5$\pm0.1$ & 78.5$\pm1.2$ & 76.4$\pm0.7$ & 62.8$\pm1.3$ & 61.7$\pm1.6$ & 72.1 \\
4. & \cmark & \xmark & \cmark & \xmark & 82.4$\pm0.9$ & 74.4$\pm0.8$ & 75.1$\pm0.9$ & 73.7$\pm0.1$ & 80.2$\pm0.9$ & 78.5$\pm0.5$ & 65.0$\pm0.9$ & 65.0$\pm1.4$ & 74.3 \\
5. & \cmark & \xmark & \cmark & \cmark & 82.6$\pm0.6$ & 74.7$\pm0.3$ & 75.9$\pm0.4$ & 73.9$\pm0.4$ & 81.4$\pm0.8$ & 78.3$\pm0.9$ & 64.5$\pm0.9$ & 66.4$\pm1.4$ & 74.7 \\
\end{tabular}
\caption{Averaged accuracy scores over 8 languages for UD POS tagging.}
\label{tab:pos}
\end{table*}

\begin{table*}[ht]
\setlength\tabcolsep{4pt}
\small
\centering
\begin{tabular}{l|cccc||cccccccc|c}
\hlineB{4}
& \multicolumn{4}{c||}{\bf \textsc{Embeddings}} & \multicolumn{9}{c}{\bf BiLSTM-CRF models on UD POS tagging} \\  
\hhline{~|----||---------}
 &  \textbf{M} & \textbf{F} & \textbf{W} & \textbf{C}  & ar & cs & de & en & es & fr & nl & ta & Avg. \\
\hline\hline
\multicolumn{13}{c}{\bf \textsc{Low-Resource: 10 Sentences}}\\
\hline
1. & \xmark & \cmark & \xmark & \xmark & 85.5$\pm0.3$ & 81.6$\pm0.8$ & \textbf{82.9}$\pm0.2$ & 77.1$\pm0.0$ & 87.9$\pm0.3$ & 84.9$\pm0.7$ & \textbf{70.7}$\pm0.7$ & 71.5$\pm0.8$ & 80.2 \\
2. & \xmark & \cmark & \cmark & \xmark & \textbf{86.3}$\pm0.5$ & \textbf{82.0}$\pm0.2$ & 82.5$\pm0.4$ & \textbf{78.9}$\pm0.4$ & \textbf{88.2}$\pm0.3$ & \textbf{85.2}$\pm0.4$ & 70.5$\pm1.1$ & 72.6$\pm0.5$ & \textbf{80.8} \\
3. & \cmark & \xmark & \xmark & \xmark & 79.7$\pm1.2$ & 72.3$\pm0.4$ & 72.6$\pm1.0$ & 69.0$\pm0.3$ & 78.8$\pm0.6$ & 76.7$\pm0.5$ & 62.5$\pm1.4$ & 62.3$\pm0.8$ & 71.7 \\
4. & \cmark & \xmark & \cmark & \xmark & 81.7$\pm0.4$ & 74.2$\pm0.8$ & 74.2$\pm0.8$ & 70.9$\pm0.4$ & 80.7$\pm0.3$ & 77.8$\pm0.7$ & 62.7$\pm0.7$ & 66.4$\pm1.2$ & 73.6 \\
5. & \cmark & \xmark & \cmark & \cmark & 82.0$\pm0.5$ & 74.0$\pm0.8$ & 73.9$\pm0.8$ & 70.6$\pm1.9$ & 81.0$\pm0.4$ & 78.5$\pm0.5$ & 63.6$\pm1.2$ & 66.4$\pm1.2$ & 73.7 \\
\end{tabular}
\caption{Averaged accuracy scores over 8 languages for UD POS tagging with the CRF layer.}
\label{tab:crf_pos}
\end{table*}

\begin{table}[]
\small
\centering
\begin{tabular}{l|cccc||cc|c}
\hlineB{4}
& \multicolumn{4}{c||}{\bf \textsc{Embeddings}} & \multicolumn{3}{c}{\bf MaxEnt models on chunking} \\  
\hhline{~|----||---}
 &  \textbf{M} & \textbf{F} & \textbf{W} & \textbf{C}  & de & en & Avg. \\
\hline\hline
\multicolumn{8}{c}{\bf \textsc{Low-Resource: 10 Sentences}}\\
\hline
1. & \xmark & \cmark & \xmark & \xmark & \textbf{83.3}$\pm0.6$ & 71.8$\pm0.3$ & 77.6 \\
2. & \xmark & \cmark & \cmark & \xmark & 83.2$\pm0.6$ & \textbf{73.3}$\pm0.6$ & \textbf{78.2} \\
3. & \cmark & \xmark & \xmark & \xmark & 64.9$\pm1.1$ & 65.1$\pm1.0$ & 65.0 \\
4. & \cmark & \xmark & \cmark & \xmark & 66.3$\pm3.0$ & 67.7$\pm0.6$ & 67.0 \\
5. & \cmark & \xmark & \cmark & \cmark & 66.5$\pm1.8$ & 67.0$\pm0.3$ & 66.7 \\
\end{tabular}
\caption{Averaged F1 scores over 2 languages for chunking.}
\label{tab:chunk}
\end{table}

\begin{table}[]
\small
\centering
\begin{tabular}{l|cccc||cc|c}
\hlineB{4}
& \multicolumn{4}{c||}{\bf \textsc{Embeddings}} & \multicolumn{3}{c}{\bf CRF models on chunking} \\  
\hhline{~|----||---}
 &  \textbf{M} & \textbf{F} & \textbf{W} & \textbf{C}  & de & en & Avg. \\
\hline\hline
\multicolumn{8}{c}{\bf \textsc{Low-Resource: 10 Sentences}}\\
\hline
1. & \xmark & \cmark & \xmark & \xmark & \textbf{83.0}$\pm0.8$ & 70.1$\pm0.1$ & \textbf{76.5} \\
2. & \xmark & \cmark & \cmark & \xmark & 82.6$\pm1.1$ & \textbf{70.5}$\pm0.1$ & \textbf{76.5} \\
3. & \cmark & \xmark & \xmark & \xmark & 68.0$\pm2.1$ & 62.9$\pm1.2$ & 65.4 \\
4. & \cmark & \xmark & \cmark & \xmark & 71.2$\pm1.8$ & 62.1$\pm0.6$ & 66.6 \\
5. & \cmark & \xmark & \cmark & \cmark & 71.0$\pm1.0$ & 61.2$\pm0.8$ & 66.1 \\
\end{tabular}
\caption{Averaged F1 scores over 2 languages for chunking with the CRF layer.}
\label{tab:crf_chunk}
\end{table}

\begin{table*}[]
\setlength\tabcolsep{4pt}
\small
\centering
\begin{tabular}{ccccc||cccccccc|c}
\hlineB{4}
\multicolumn{5}{c||}{\bf \textsc{Embeddings}} & \multicolumn{9}{c}{\bf Languages} \\  
 \hline\hline
\multirow{2}{*}{\textbf{M}} & \multirow{2}{*}{\textbf{F}} & \multirow{2}{*}{\textbf{W}} & \multirow{2}{*}{\textbf{C}} & \multirow{2}{*}{\textbf{MF}} & \multicolumn{9}{c}{\bf \textsc{NER}}\\
\hhline{~~~~~||--------|-}
 &  &  &  &  & ar & cs & de & en & es & fr & nl & ta & Avg. \\
 \hline
\cmark & \cmark & \cmark & \cmark & \xmark & 84.6$\pm0.2$ & 88.4$\pm0.2$ & \textbf{85.5}$\pm0.3$ & 81.7$\pm0.2$ & \textbf{90.6}$\pm0.2$ & \textbf{88.4}$\pm0.5$ & \textbf{90.1}$\pm0.3$ & \textbf{85.5}$\pm0.3$ & \textbf{86.8} \\
\cmark & \xmark & \cmark & \cmark & \cmark & 83.6$\pm0.8$ & 87.7$\pm0.3$ & 84.2$\pm0.3$ & 81.4$\pm0.1$ & 90.0$\pm0.1$ & 87.9$\pm0.1$ & 89.9$\pm0.4$ & 84.2$\pm0.2$ & 86.1 \\
\cmark & \cmark & \cmark & \cmark & \cmark & \textbf{84.8}$\pm0.5$ & \textbf{88.6}$\pm0.2$ & 85.1$\pm0.3$ & \textbf{82.0}$\pm0.2$ & 90.3$\pm0.4$ & 88.1$\pm0.3$ & \textbf{90.1}$\pm0.2$ & 85.3$\pm0.3$ & \textbf{86.8} \\
\hline\hline
& & & & & \multicolumn{9}{c}{\bf \textsc{POS Tagging}}\\
\hhline{~~~~~||--------|-}
 &  &  &  &  &  ar & cs & de & en & es & fr & nl & ta & Avg. \\
\end{tabular}
\caption{Detailed comparison for additionally concatenating \textbf{MF} with \textbf{All}. \textbf{MF} represents the M-Flair embeddings.}
\label{tab:mflair}
\end{table*}

\end{document}